\documentclass[letterpaper, 10 pt, conference]{ieeeconf}  

\IEEEoverridecommandlockouts                              

\usepackage{newtxtext,newtxmath}
\usepackage{algorithm}
\usepackage[noend]{algpseudocode}
\usepackage{tabularx,ragged2e,booktabs}
\usepackage{tabulary}
\usepackage{bm}
\usepackage{cite}
\usepackage{soul}

\usepackage{graphics} 
\usepackage{epsfig} 
\usepackage{mathtools}
\usepackage{times} 
\usepackage{xcolor}
\usepackage{multicol}
\usepackage{multirow}
\usepackage{colortbl}
\usepackage{lipsum}

\title{\LARGE \bf
MOCA-S: A Sensitive Mobile Collaborative Robotic Assistant exploiting Low-Cost Capacitive Tactile Cover and Whole-Body Control
}

\author{Mattia Leonori$^*$, Juan M. Gandarias$^*$~\IEEEmembership{Member, IEEE}, and Arash Ajoudani~\IEEEmembership{Member, IEEE}
\thanks{This work was supported in part by the European Union’s Horizon 2020 research and innovation programme under Grant Agreement No. 871237 (SOPHIA) in part by the ERC-StG Ergo-Lean (Grant Agreement No.850932).}
\thanks{The authors are with HRI$^{2}$ Lab, Istituto Italiano di Tecnologia, 16163 Genoa, Italy
        {\tt\small \{mattia.leonori, juan.gandarias, arash.ajoudani\}@iit.it}
}
\thanks{$^*$Contributed equally to the work.}
}

\begin{document}

\maketitle
\thispagestyle{empty}
\pagestyle{empty}

\begin{abstract}
Safety is one of the most fundamental aspects of robotics, especially when it comes to collaborative robots (cobots) that are expected to physically interact with humans. Although a large body of literature has focused on safety-related aspects for fixed-based cobots, a low effort has been put into developing collaborative mobile manipulators. In response to this need, this work presents MOCA-S, i.e., Sensitive Mobile Collaborative Robotic Assistant, that integrates a low-cost, capacitive tactile cover to measure interaction forces applied to the robot base. The tactile cover comprises a set of 11 capacitive large-area tactile sensors distributed as a 1-D tactile array around the base. Characterization of the tactile sensors with different materials is included. 
Moreover, two expanded whole-body controllers that exploit the platform's tactile cover and the loco-manipulation features are proposed. These controllers are tested in two experiments, demonstrating the potential of MOCA-S for safe physical Human-Robot Interaction (pHRI). Finally, an experiment is carried out in which an undesired collision occurs between MOCA-S and a human during a loco-manipulation task. The results demonstrate the intrinsic safety of MOCA-S and the proposed controllers, suggesting a new step towards creating safe mobile manipulators.
\end{abstract}

\section{Introduction}
\label{sec:intro}

\begin{figure}
    \centering
    \includegraphics[width=1\columnwidth]{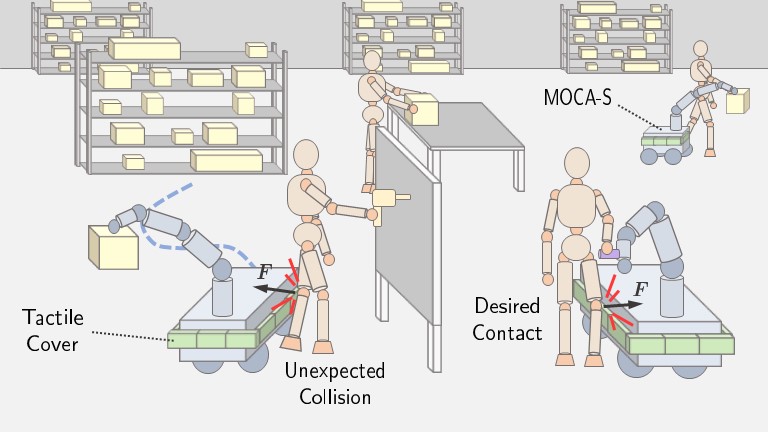}
    \caption{Illustration of a workplace with multiple collaborative mobile manipulators where both unexpected and desired physical contacts with the robot base may occur when humans and robots are interactively working in close contact.}
    \label{fig:digest}
\end{figure}

Safety in robotics is one of the most critical aspects, especially concerning Human-Robot Interaction (HRI). It becomes even more significant in industrial environments, where safety requirements are elevated. In this regard, the emergence of collaborative robots (cobots) in logistics and manufacturing has seen significant advances in terms of safety, even from the legislation perspective (e.g., ISO/TS 15066~\cite{rosenstrauch2017safe, ferraguti2020control}) allowing close collaborations between robots and humans~\cite{heo2019collision, weiss2021cobots} (see Fig.~\ref{fig:digest}). 
Nevertheless, most of these developments focus on fixed-based robots. In the case of mobile manipulators or floating-based robots, the number of contributions in the literature is much less significant. Most existing systems, especially the more advanced ones, focus on vision systems or lasers together with artificial intelligence algorithms~\cite{chung2011detection, fotiadis2013human, gupta2016novel}. However, these systems might fail based on different sources (e.g., occlusions, reflection, range detection). It is well-known that these systems are not yet robust enough to work in close proximity with humans~\cite{dietterich2017steps, ren2018learning, rice2020overfitting}, hence maybe be insufficient for robot safety legislation.

Different techniques have been proposed with robotic manipulators to ensure safety when physical interactions with humans occur. Impedance-based controllers are one of the most studied approaches~\cite{fasse1997spatial, albu2004passivity, de2006collision}. However, soft robots~\cite{gandarias2020open} or the use of tactile sensors~\cite{fritzsche2011tactile, gandarias2018enhancing, pang2020coboskin} have been also proposed. Despite this, little effort has been put into developing intrinsically safe floating-based robots. Although some works have incorporated whole-body Cartesian impedance controllers to render safe and compliant robot behaviors~\cite{dietrich2016whole, bussmann2018whole, wu2021unified}, however, the control of physical interaction mainly focused on the arm and not the mobile bases. The most significant work is found in the development of the robot TOMM~\cite{dean2017tomm}, which has tactile sensors distributed throughout the upper body of the robot. However, this robot does not have tactile sensors distributed throughout the base, hence, there is the possibility of an accident if the base collides with a person. 

This work tackles the aforementioned problem and proposes a solution based on a tactile cover integrated on the base of a Mobile Collaborative Robotic Assistant (MOCA). The cover encloses a one-dimensional array of capacitive tactile sensors distributed around the robot's base. An experiment with different dielectric materials is performed in order to choose the most suitable sensor for the addressed problem. In addition, the extension and development of two whole-body controllers exploiting the benefits of the tactile cover are discussed. Finally, experiments are performed to analyze the behavior and performance of the robot and the controllers under different circumstances. Therefore, the contributions of this work can be listed as follows:

\begin{figure*}
    \centering
    \includegraphics[width=0.8\textwidth]{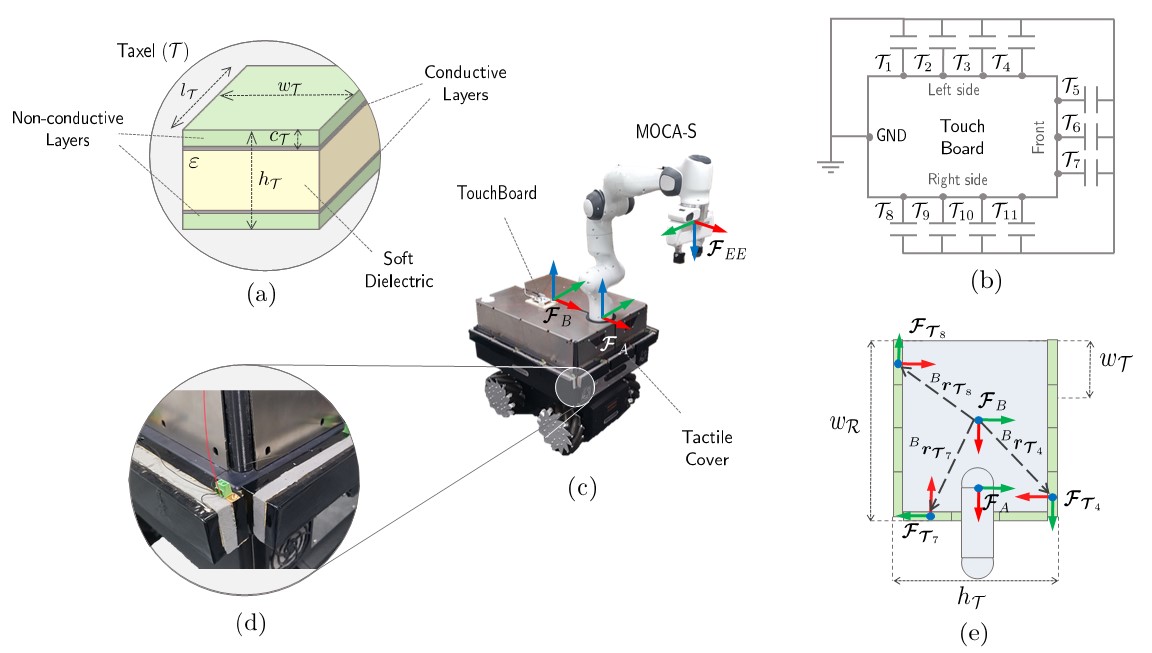}
    \caption{Composition and integration of the tactile cover on the MOCA-S platform. (a) An illustration of the components and parameters of a capacitive tactile unit (taxel). (b) A schematic view of the electronics. (c) A complete view of the MOCA-S prototype platform with the \textit{Touch Board} and the tactile cover. (d) A detailed view of the large-area taxel prototypes and the connection of the terminals of the capacitors. (e) A representative of the tactile cover's integration and how the taxels are distributed around the platform.}
    \label{fig:tactile_sensor}
\end{figure*}

\begin{itemize}
    \item The presentation of MOCA-S, a sensitive mobile manipulator that extends the concept of our robot MOCA by integrating a low-cost tactile cover formed by capacitive tactile sensors.
    \item An study and analysis for the characterization and calibration of the capacitive tactile sensors using different materials. 
    \item The proposal, implementation, and experimentation of two extended whole-body controllers to demonstrate the potential and benefits of MOCA-S in close human-robot collaboration environments.
    \item A safety experiment involving an unexpected collision of the mobile base with a human demonstrating the intrinsic safety of MOCA-S even if unintended interactions occur.
\end{itemize}

This paper is structured as follows: Section~\ref{sec:MOCA-S} describes the MOCA-S platform, including a description of the proposed capacitive tactile sensor, and the integration of the tactile cover on the robot. Section~\ref{sec:controllers} presents the two expanded whole-body controllers from a theoretical point of view. 
The experiments and results are included in section~\ref{sec:experiments}. Finally, the conclusions of the work are presented in section~\ref{sec:conclusions}.

\section{MOCA-S: Sensitive Mobile Collaborative Robotic Assistant}
\label{sec:MOCA-S}

This section describes the MOCA-S platform, a sensitive mobile manipulator that extends the concept of our previous robot, Mobile Collaborative Robotic Assistant (MOCA), already employed in multiple works~\cite{wu2019teleoperation, kim2020moca, ruiz2022improving, gandarias2022enhancing}. The MOCA robot comprises a \textit{Robotnik SUMMIT-XL STEEL} mobile platform and a 7 DoFs, torque-controlled \textit{Franka Emika Panda} manipulator, while MOCA-S also integrates a tactile cover as described below.

\subsection{Large-area Capacitive Tactile Sensor}
\label{subsec:tactile_cover}

The tactile cover consists of 11 tactile sensor units, also called taxels or tactels, distributed around the mobile platform. Each taxel is a parallel-plate capacitor consisting of 5 layers, as illustrated in Fig.~\ref{fig:tactile_sensor}a. These layers are distributed as follows:

\begin{enumerate}
    \item Layer 1: The first layer is composed of a rigid non-conductive material. 
    \item Layer 2: A layer of conductive material forms the first terminal of the capacitor sensor. This terminal is connected to the ground, as explained in more detail in section~\ref{subsec:touch_board}.
    \item Layer 3: The third layer comprises a soft dielectric material. A detailed experiment with different materials and sensor characterization and calibration is included in section~\ref{subsec:sensor_characterization}.
    \item Layer 4: Again, another layer of conductive material constitutes the second terminal of the capacitor, which is connected to an electrode of the \textit{Touch Board} (see section~\ref{subsec:touch_board} for more details).
    \item Layer 5: The last one is formed by another layer of rigid non-conductive material.
\end{enumerate}

Each taxel $\mathcal{T}_i$ is a tactile sensor unit that can measure the Capacitance based on the following equation
\begin{equation}
    C_i = \varepsilon \frac{A}{d_i},
\end{equation}
where $C_i$ is the Capacitance, $\varepsilon$ is the permittivity of the dielectric, $A$ is the surface of the terminal, and $d_i$ is the distance between the terminals. Here we assume that the $\varepsilon$ and $A$ are constant and have the same value for all taxels. Hence, the only parameter that can vary the system's Capacitance is $d_i$. Therefore, a soft dielectric is used to vary this distance when a compressing force is applied to the external terminal. The fact that the dielectric is deformable also promotes safety as they produce soft contacts. The calibration of the sensor, i.e., how to transform the capacitance lectures into forces, is described in section~\ref{subsec:sensor_characterization}.

\subsection{Electronics: \textit{Touch Board} Microcontroller}
\label{subsec:touch_board}
The \textit{Touch Board} \footnote{https://www.bareconductive.com/products/touch-board} (Bare Conductive, UK) is used to read and process the data given by the tactile sensor.
The \textit{Touch Board} is a microcontroller based on the \textit{ATmega32U4} microprocessor that integrates the dedicated capacitive touch sensor driver \textit{MPR121}. This driver allows the lecture of up to 12 capacitive touch electrodes. 

A schematic view of the connection between the capacitors and the \textit{Touch Board} is shown in~\ref{fig:tactile_sensor}b. Each taxel is composed of a parallel-plate capacitor that has one of the terminals (fixed terminal) connected to one board electrode, while the other (moving terminal) is connected to the board's ground and the moving terminals of the other taxels.

\subsection{Tactile Cover Integration}
\label{subsec:cover_integration}

The integration of the tactile cover on the platform is depicted in Fig.~\ref{fig:tactile_sensor}c, with a detailed picture of the taxels connection in~\ref{fig:tactile_sensor}d. This figure also shows how the connection of the terminals of the taxels is made. 
The cover is a one-dimensional tactile array that can measure the force distribution around the longitudinal to the platform.

The distribution of the 11 taxels around the platform is shown in~\ref{fig:tactile_sensor}e. This distribution is carried out assuming that the robot's interaction with the environment will occur in the proximity of the robot end-effector; therefore, the taxels are distributed only on the robot's front and right and left sides. Nevertheless, if a particular application would require the integration of the sensors on the robot's backside for any specific reason, the taxels can be distributed. Moreover, considering the differences in lengths between the front and the robot's sides, we decided to put four taxels on the sides and three on the front. 
A noteworthy aspect is the relative pose of the sensor frame with respect to the base frame. This aspect is considered in the next section to transform the forces measured by the sensors to the robot base frame.

\begin{figure}
    \centering
    \includegraphics[width=0.65\columnwidth]{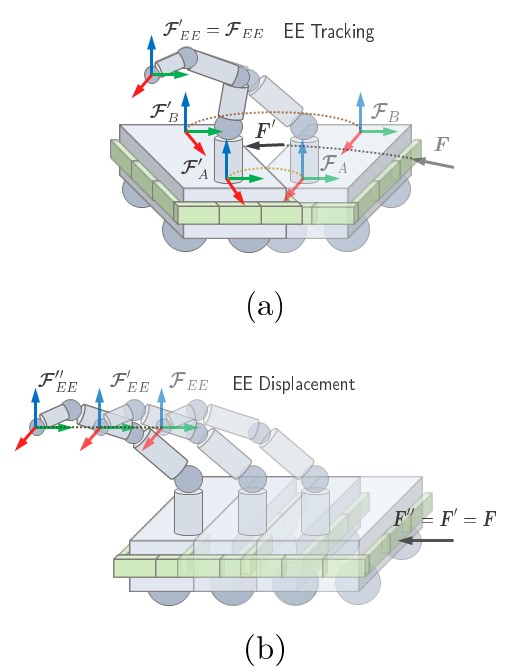}   \caption{Behavior of the expanded controllers when an external force is applied to the robot base. (a) Expanded Whole-Body Cartesian Impedance Controller: the base moves while the position of the end-effector remains, exhibiting a compliance behavior at the Cartesian level. (b) Expanded Whole-Body Admittance Controller: Both the end-effector and the base move according to a whole-body admittance control law.}
    \label{fig:expanded_controllers}
\end{figure}

\section{Expanded Whole-Body Controllers}
\label{sec:controllers}

Two whole-body controllers that exploit the advantages provided by the tactile cover are proposed below.

\subsection{Expanded Whole-Body Cartesian Impedance Controller}
\label{subsec:wb_controller}

The first controller is an expanded whole-body Cartesian impedance controller that builds on our previous controller presented in~\cite{wu2021unified}. An illustration of the behavior of this controller when an external force is applied to the base is depicted in Fig.~\ref{fig:expanded_controllers}a. 
The dynamics of the robot are defined as a torque-controlled floating base system. As the base is velocity-controlled, the following admittance control law is required to convert virtual control torques into desired velocities
\begin{equation}
\label{eq:admittance_eq_base}
    \boldsymbol{M}_{v}\boldsymbol{\ddot{q}}_{B} + \boldsymbol{D}_{v}\boldsymbol{\dot{q}}_{B} = \boldsymbol{\tau}_{v},
\end{equation}
where $\boldsymbol{M}_{v}, \boldsymbol{D}_{v} \in \mathbb{R}^{n_B \times n_B}$ are the diagonal positive definite virtual mass and
damping matrices of the base; $\boldsymbol{q}_B, {\boldsymbol{\dot{q}}}_B,{\boldsymbol{\ddot{q}}}_B \in \mathbb{R}^{n_B}$ are the current mobile base joint positions, velocities and accelerations; and $\boldsymbol{\tau}_{v} \in \mathbb{R}^{3}$ are the virtual base torques vector, being $n_B$ the number of degrees of freedom of the base (in the MOCA-S particular case $n_B=3$). Hence, the dynamics of the platform are defined by
\begin{equation}
\begin{aligned}
\label{eq:whole_body_dynamics}
&  \begin{bmatrix}
 \boldsymbol{M}_v & \boldsymbol{0} \\
 \boldsymbol{0} & \boldsymbol{M}_A(\boldsymbol{q}_A)
\end{bmatrix} \boldsymbol{\ddot{q}} +
\begin{bmatrix}
 \boldsymbol{D}_v & \boldsymbol{0} \\
 \boldsymbol{0} & \boldsymbol{C}_A(\boldsymbol{q}_A,\dot{\boldsymbol{q}}_A)
\end{bmatrix} \boldsymbol{\dot{q}} \\
& + \begin{bmatrix}
 \boldsymbol{0} \\
 \boldsymbol{g}_A(\boldsymbol{q}_A)
\end{bmatrix} = \begin{bmatrix} \boldsymbol{\tau}_v^T \\ \boldsymbol{\tau}_A^T \end{bmatrix} + \begin{bmatrix} \boldsymbol{\tau}_{v,ext}^T \\ \boldsymbol{\tau}_{A,ext}^T \end{bmatrix},
\end{aligned}
\end{equation}
where $\boldsymbol{M}_A(\boldsymbol{q}_A)\in\mathbb{R}^{n_A \times n_A}, \, \boldsymbol{C}_A(\boldsymbol{q}_A,\boldsymbol{\dot{q}}_A)\in\mathbb{R}^{n_A \times n_A}, \, \boldsymbol{g}_A(\boldsymbol{q}_A)\in\mathbb{R}^{n_A}$ are the arm mass, the Coriolis and centrifugal terms matrices, and the gravity vector, respectively; $\boldsymbol{q}, {\boldsymbol{\dot{q}}},{\boldsymbol{\ddot{q}}}\in\mathbb{R}^{n}$ are the current whole-body joint positions, velocities and accelerations; $\boldsymbol{\tau}_{A,ext}, \boldsymbol{\tau}_{A} \in \mathbb{R}^{n_A}$ are the arm external and control torque vectors; being $n_A$ and $n$ the number of degrees of freedom of the arm and whole platform (in the MOCA-S particular case, $n_A=7$ and $n=10$). May the reader note that the difference with~\cite{wu2021unified} is the presence of the virtual base external torques vector $\boldsymbol{\tau}_{v,ext}^T\in\mathbb{R}^{n_B}$, that are computed as
\definecolor{myorange}{rgb}{1,0.37,0.12}
\begin{equation}
    \label{eq:virtual_torques_base_wb}
    \boldsymbol{\tau}_{v,ext}^T = \begin{bmatrix} \sum_{i=1}^{n_\mathcal{T}} {^B}\boldsymbol{R}_{\mathcal{T}_i} \begin{bmatrix}F_{\mathcal{T}_i} \\ 0 \end{bmatrix} \vspace{0.2cm} \\
    \sum_{i=1}^{n_\mathcal{T}} {^B}\boldsymbol{r}_{\mathcal{T}_i} \times {^B}\boldsymbol{R}_{\mathcal{T}_i} \begin{bmatrix}F_{\mathcal{T}_i} \\ 0 \end{bmatrix} \end{bmatrix},
\end{equation}
where $n_\mathcal{T}$ is the number of taxels; ${^B}\boldsymbol{r}_{\mathcal{T}_i}\in\mathbb{R}^{2}$, ${^B}\boldsymbol{R}_{\mathcal{T}_i}\in\mathbb{R}^{2 \times 2}$ and  are, respectively, the position, the rotation matrix of the $i$-th sensor with respect to the mobile base frame $\boldsymbol{\mathcal{F}}_B$ (see Fig.~\ref{fig:tactile_sensor}e); and ${F}_{\mathcal{T}_i}\in\mathbb{R}$ is the sensed interaction force on the $x$-axis of the $i$-th sensor. The rotation matrix of a particular taxel $i$ is given by
\begin{equation}
    \label{eq:rotation_matrix}
    {^B}\boldsymbol{R}_{\mathcal{T}_i} = \begin{bmatrix}\cos{\phi_{\mathcal{T}_i}} & -\sin{\phi_{\mathcal{T}_i}}\\ \sin{\phi_{\mathcal{T}_i}} & \cos{\phi_{\mathcal{T}_i}} \end{bmatrix},
\end{equation}
where $\phi_{\mathcal{T}_i}\in\mathbb{R}$ is the orientation of the sensor with respect to $\boldsymbol{\mathcal{F}}_B$. Note that this matrix considers a rotation around the Z-axis of $\boldsymbol{\mathcal{F}}_B$.

The whole-body controller generates torque references $\boldsymbol{\tau}_c = \begin{bmatrix} \boldsymbol{\tau}_v^T & \boldsymbol{\tau}_A^T \end{bmatrix}^T \in \mathbb{R}^n$ that are passed to the admittance controller of the base (defined in equation~(\ref{eq:admittance_eq_base})) and to the arm. The relationship between $\boldsymbol{\tau}_c$ and the generalized Cartesian forces $\boldsymbol{F} \in \mathbb{R}^6$ is defined by the following control law (note that for the sake of readability, the dependencies are drop.)
\begin{equation}
\label{eq:opt_solution}
\begin{aligned}
    \boldsymbol{\tau}_c & = \boldsymbol{W^{-1}M^{-1}J^{T}\Lambda_{W}\Lambda^{-1}F}\\
    & + (\boldsymbol{I-W^{-1}M^{-1}J^{T}\Lambda_{W}JM^{-1}})\boldsymbol{\tau}_0 ,
\end{aligned}
\end{equation}
representing the desired impedance behavior $\boldsymbol{\bar{J}}^T \boldsymbol{\tau}_c = \boldsymbol{F} $, where
\begin{align}
    \notag \bar{\boldsymbol{J}} &= \boldsymbol{M}^{-1} \boldsymbol{J}^{T} \boldsymbol{\Lambda}  \; ,\\
    \boldsymbol{\Lambda_{W}} &= \boldsymbol{J}^{-T}\boldsymbol{MWM}\boldsymbol{J}^{-1}  \; ,\\
    \notag \boldsymbol{\Lambda} &= {\left( \boldsymbol{J}\boldsymbol{M}^{-1} \boldsymbol{J}^{T} \right)}^{-1}  \; ,
\end{align}
where $\boldsymbol{\Lambda}, \boldsymbol{\Lambda_{W}} \in \mathbb{R}^{6 \times 6}$ are the unweighted and weighted Cartesian Inertia, respectively; $\boldsymbol{J}, \bar{\boldsymbol{J}} \in \mathbb{R}^{6 \times n}$ are the whole-body geometric Jacobian and the dynamically consistent Jacobian, respectively; $\boldsymbol{I}\in \mathbb{R}^{n \times n}$ is the identity matrix.

The null-space torque $\boldsymbol{\tau}_0 \in \mathbb{R}^n$ is used to generate motions that do not interfere with the Cartesian force $\boldsymbol{F}$.

The positive definite weighting matrix $\boldsymbol{W} \in \mathbb{R}^{n \times n}$ is defined as
\begin{equation} 
\label{eq:weight}  
    \boldsymbol{W}(\boldsymbol{q})=\boldsymbol{H}^T\boldsymbol{M}^{-1}(\boldsymbol{q})\boldsymbol{H},
\end{equation}
where $\boldsymbol{H} \in \mathbb{R}^{n \times n}$ is the diagonal positive definite controller weighting matrix that is dynamically selected based on the loco-manipulation gains $\eta_A, \eta_B \in \mathbb{R}_{>0}$ according to the following equation
\begin{equation} 
\label{eq:eta}
    \boldsymbol{H} = \begin{bmatrix} \eta_{B}\boldsymbol{I}_{n_B} & \boldsymbol{0}_{n_B\times n_B} \\ \boldsymbol{0}_{n_B\times n_B} & \eta_{A} \boldsymbol{I}_{n_B} \end{bmatrix},
\end{equation}

Hence, the desired Cartesian impedance behavior is obtained by
\begin{equation}
\boldsymbol{F} =   \boldsymbol{D}_d(\dot{\boldsymbol{x}}_d-\dot{\boldsymbol{x}}) + \boldsymbol{K}_d(\boldsymbol{x}_d - \boldsymbol{x}),
    \label{eq:cartesian_impedance}
\end{equation}
where $\boldsymbol{x}, \boldsymbol{\dot{x}} \in \mathbb{R}^{6}$ are the current end-effector pose and twist; and $\boldsymbol{D}_d, \boldsymbol{K}_d \in \mathbb{R}^{6 \times 6}$ are the desired Cartesian damping and stiffness. 

Finally, the null-space torque $\boldsymbol{\tau}_0$ is generated as
\begin{equation} 
    \label{eq:nullspace_impedance}
    \boldsymbol{\tau}_{0} = \begin{bmatrix} \boldsymbol{\tau}_{v,ext}^T \\ -\boldsymbol{D}_0\dot{\boldsymbol{q}_A} + \boldsymbol{K}_0(\boldsymbol{q}_{A, ref} - \boldsymbol{q_A})
    \end{bmatrix}.
\end{equation}
In contrast to~\cite{wu2021unified}, the virtual base external torques
vector are considered in the null-space torque generation. Besides, the value of the null-space damping and stiffness matrices of the arm $\left(\boldsymbol{D}_0, \boldsymbol{K}_0 \in \mathbb{R}^{n_A \times n_A}\right)$ are important to define the loco-manipulation behavior of the robot.

\subsection{Expanded Follow Me: Whole-Body Admittance Controller}
\label{subsec:follow_me}
The second controller is an expanded whole-body admittance controller that builds on the haptic follow-me controller of our previous work presented in~\cite{lamon2020visuo}. An illustration of the behavior of the robot thanks to this controller when an external force is applied to the base is shown in Fig.~\ref{fig:expanded_controllers}b. 

The whole-body dynamics of the system and the control law are the same as in the whole-body impedance controller as previously defined in equations~(\ref{eq:whole_body_dynamics}). However, in this case, the external wrenched applied at the end-effector are also considered in equation~(\ref{eq:virtual_torques_base_wb}). Hence,
%
\begin{equation}
    \label{eq:virtual_torques_base_follow_me}
    \boldsymbol{\tau}_{v,ext}^T = \begin{bmatrix} \sum_{i=1}^{n_\mathcal{T}} {^B}\boldsymbol{R}_{\mathcal{T}_i} \begin{bmatrix}F_{\mathcal{T}_i} \\ 0 \end{bmatrix} \vspace{0.2cm} \\
    \sum_{i=1}^{n_\mathcal{T}} {^B}\boldsymbol{r}_{\mathcal{T}_i} \times {^B}\boldsymbol{R}_{\mathcal{T}_i} \begin{bmatrix}F_{\mathcal{T}_i} \\ 0 \end{bmatrix} \end{bmatrix}+{^B}\boldsymbol{T}_{A}\boldsymbol{J}_{A}^T\boldsymbol{\tau}_{A,ext}^T
\end{equation}
where ${^B}\boldsymbol{T}_{A}\in\mathbb{R}^{3 \times 3}$ is the transformation matrix among generalized forces applied at the end-effector w.r.t. the arm base frame $\boldsymbol{\mathcal{F}}_A$ transformed in mobile base frame $\boldsymbol{\mathcal{F}}_B$; and $\boldsymbol{J}_{A}^T\in\mathbb{R}^{n_A \times 6}$ is the transpose Jacobian of the arm. Hence, in this controller, external forces applied at the base (i.e., first component of equation~(\ref{eq:virtual_torques_base_follow_me})) or external forces applied at the arm (i.e., second component of equation~(\ref{eq:virtual_torques_base_follow_me})) cause a motion of the base. The transformation matrix between the base and the arm is given by
\begin{equation}
    ^B\boldsymbol{T}_{A} = \begin{bmatrix} {^B}\boldsymbol{R}_{A} & \boldsymbol{0}_{2 \times 2}\\-\boldsymbol{S}^{T}({^B}\boldsymbol{r}_{BA}){^B}\boldsymbol{R}_{A} & 1
    \end{bmatrix}
\end{equation}
where $\boldsymbol{S}^{T}({^B}\boldsymbol{r}_{BA})$ is the skew-symmetric matrix of the position of the arm base link in $\boldsymbol{\mathcal{F}}_B$.

As in the case of the previous controller, a two-level priority Cartesian torque control is exploited as
\begin{equation}
\label{eq:solution_follow_me}
    \boldsymbol{\tau}_c = \boldsymbol{J}_A^T\boldsymbol{F} + \left(\boldsymbol{I}-\boldsymbol{J}_A^T\boldsymbol{\Lambda}\boldsymbol{J}_A\boldsymbol{M}_A^{-1}\right)\boldsymbol{\tau}_0.
\end{equation}
Moreover, the null-space torque $\boldsymbol{\tau}_0$ previously calculated in equation~(\ref{eq:nullspace_impedance}) is now computed as
\begin{equation} 
    \label{eq:nullspace_impedance_follow_me}
    \boldsymbol{\tau}_{0} = \left[ -\boldsymbol{D}_0\dot{\boldsymbol{q}_A} + \boldsymbol{K}_0(\boldsymbol{q}_{A, ref} - \boldsymbol{q_A}) \right].
\end{equation}
Therefore, in this case, as the null-space torque is not considering $\boldsymbol{\tau}_{v,ext}$, the second task of the controller contributes on keeping the initial configuration of the arm $\boldsymbol{q}_{A, ref}$.

\section{Experiments and Results}
\label{sec:experiments}

This section includes the results of this work, splitted in three experiments that are described below. A video~\footnote{https://youtu.be/IX6fn8ODSt8} with the complete realization of the experiments is uploaded as supplementary material. Regarding the experiments with humans, the whole experimental procedure was carried out at Human-Robot Interfaces and Physical Interaction (HRI$^{2}$) Lab, Istituto Italiano di Tecnologia, Genoa, Italy, in accordance with the Declaration of Helsinki, and the protocol was approved by the ethics committee Azienda Sanitaria Locale (ASL) Genovese N.3 (Protocol IIT\_HRII\_ERGOLEAN 156/2020).

\subsection{Sensor Characterization and Calibration}
\label{subsec:sensor_characterization}

\begin{figure}
    \centering
    \includegraphics[width=0.85\columnwidth]{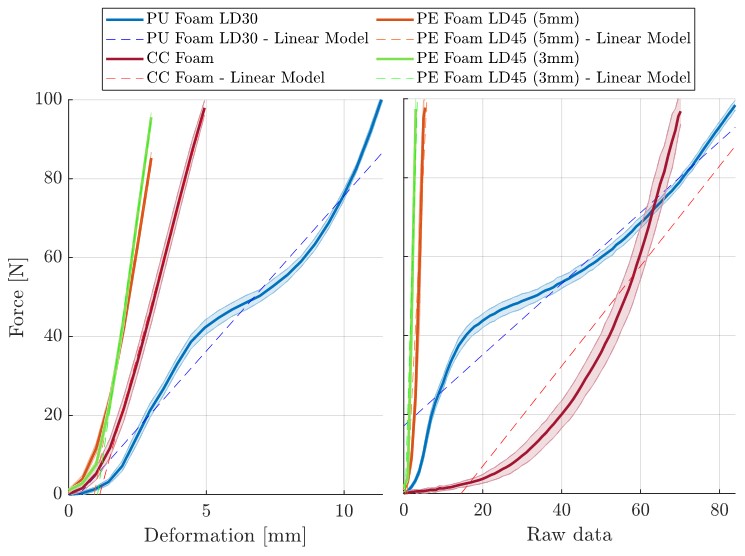}
    \caption{Results of the characterization (left) and calibration (right) experiment of one particular tactel. Each experiment is carried out ten times, resulting in the semitransparent areas representing the standard deviation, whereas the colored thin lines represent the average of each experiment.}
    \label{fig:compression_sensor}
\end{figure}

Three materials (Polyurethane (PU), polyethylene (PE) and a cellulose-cotton mix (CC) foams) are evaluated in this section as candidates for the deformable dielectric layer. In particular, one PE foam of two different thicknesses ($3mm$ and $5mm$) are included in the trials to validate the results. The experiment consists of applying a controlled compressing force while measuring the deformation and the sensed capacitance variation of one particular taxel. This test is carried out using an Universal Robot UR16 with a force/torque sensor at the end-effector. A comparison between the force and deformation allows the characterization of the materials. Moreover, a comparison between the force and the raw measurements allows the calibration of the sensor. 

The results of these comparisons are plotted in Fig.~\ref{fig:compression_sensor} and summarized in table~\ref{tab:sensor_calibration}. A lineal model is obtained for each comparison and sensor. The Root-Mean-Square-Error of each model for the characterization ($RMSE_{\Delta x}$), and the calibration ($RMSE_{\Delta s}$) are also included in the table.
According to these results, the greatest deformation ($11.4mm$), applying $100N$ is produced by the industrial polyurethane foam (PU Foam LD30), making it the most deformable material. Besides, it is the one that presents the most linear behavior according to ($RMSE_{\Delta s}$). Hence, this is the material selected for the fabrication dielectric layer of the tactile cover. The design parameters of the final taxel are listed in table~\ref{tab:sensor_design_parameters}. 

\begin{table}[]
    \centering
    \caption{Sensors Calibration and Characterization}
    \begin{tabular}{lccccc}
        \toprule
        Material &  $\Delta x[mm]$ & $\Delta s$ & $RMSE_{\Delta x}$ & $RMSE_{\Delta s}$\\
        \midrule
        \midrule
        PU Foam LD30 &  $11.4$ & $84$  & $4.064$ &  $5.96$\\
        CC Foam & $NA$ &  $70$  & $4.790$ & $11.32$ \\       
        PE Foam LD45 (5mm) &  $3.0$ & $5$  & $1.612$  & $14.37$  \\
        PE Foam LD45 (3mm) &  $2.9$ & $4$  & $1.587$  & $17.91$  \\
        \bottomrule
    \end{tabular}\\
    \flushleft
    $\Delta x$: Maximum Deformation \hspace{1cm}
    $\Delta s$: Maximum raw data variation
    \label{tab:sensor_calibration}
\end{table}

\begin{table}
    \centering
    \caption{Taxel Design Parameters}
    \begin{tabular}{cll}
        \toprule
        Symbol & Parameter & Value \\
        \midrule
        \midrule
        $h_{\mathcal{T}}$ & taxel height &  $5 mm$\\
        $w_{\mathcal{T}}$ & taxel width &   $4 mm$\\
        $l_{\mathcal{T}}$ & taxel length &  $18 mm$\\
        $c_{\mathcal{T}}$ & Rigid layer height &  $12.5 mm$\\
        $e_{\mathcal{T}}$ & Elastic layer height &  $15 mm$\\
        $d_{\mathcal{T}}$ & Elastic layer density &  $30 Kg/m^3$\\
        $c_{\mathcal{T}}$ & Rigid layer height &  $10 mm$\\
        $\mathcal{D}$ & Density &  $30Kg/m^3$\\
        $\rho$ & Conductive layer resistivity &  $2.65\times 10^{-8} \Omega \cdot m$\\
        \bottomrule
    \end{tabular}
    \label{tab:sensor_design_parameters}
\end{table}

\subsection{Expanded Controllers Demonstration}
\label{subsec:demo_controllers}

The second experiment consists of a demonstration of the robot's behavior when the aforementioned expanded controllers are applied. The experiments consist of applying multiple forces at the arm and tactile cover to analyze the robot's behavior.

\subsubsection{Expanded Impedance Controller}
\label{subsubsec:demo_wb}

\begin{figure*}
    \centering
    \includegraphics[width=0.8\textwidth]{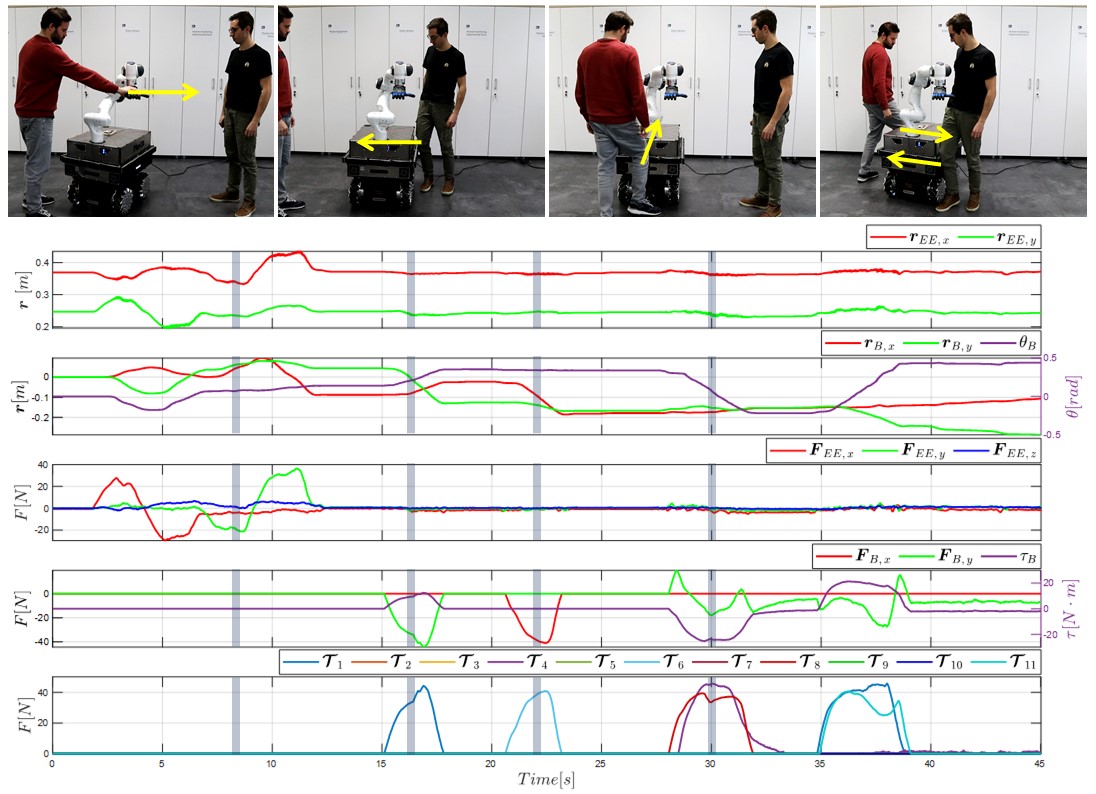}
    \caption{Demonstration of the expanded whole-body Cartesian impedance controller. From top to bottom: i) Screenshots of the experiment, where the yellow arrows define the direction of the applied forces; ii) the end-effector position (the Z component is dropped as it is constant during the experiment); iii) the pose (position--left axis, orientation--right axis) of the base; iv) the forces applied to the end-effector; v) the wrenches (force--left axis, and torque--right axis) applied to the base; vi) the forces applied to each taxel. The four semi-transparent vertical gray lines represent the moment when the four screenshots represented at the top of the figure were taken.}
    \label{fig:wb_demo}
\end{figure*}

In the case of the expanded whole-body Cartesian impedance controller, four cases are considered: i) applying forces at the arm, ii) applying forces at the side taxels, iii) applying forces at the front taxels, and iv) commanding a yaw torque at the base by applying forces at opposite side taxels.
The results of this experiment are shown in Fig~\ref{fig:wb_demo}. The four screenshots and the four semi-transparent vertical gray lines represent the previously commented cases. 

The first case demonstrates how the position of the end-effector (top plot) and the base (second plot from the top) change, exhibiting a compliance behavior according to the force read at the end-effector (third plot from the top). However, in the second case, when external forces are applied at the left side ($\mathcal{T}_1$ in the bottom plot), wrenches are read at the base (four plots from the top), causing simultaneous translation and rotation motions of the base (second plot from the top), but keeping the position of the end-effector. In the third case, when a force is applied at the front ($\mathcal{T}_6$ in the bottom plot), a pure translation is commanded to the base (second plot from the top), as this taxel is aligned with the X-axis of the base, and the end-effector remains in the same position. In the last case, when forces are applied to $\mathcal{T}_4$ and $\mathcal{T}_8$ (bottom plot), a poor rotation is commanded to the base (second plot from the top). This behavior is exhibited in the other direction ($\mathcal{T}_1$ and $\mathcal{T}_{11}$) between seconds 35 and 40 of the experiment.

\subsubsection{Expanded Follow-me Controller}
\label{subsubsec:demo_follow_me}

Regarding the expanded Follow-me controller, the following four cases are considered: i) applying forces at the arm, ii) applying forces at the left-side taxels, iii) applying forces in the same direction to the arm and the base., and iv) applying opposite forces to the arm and the base.
The results are depicted in Fig~\ref{fig:wb_demo}. As in the previous demonstration, the four screenshots and the four semi-transparent vertical gray lines represent the previously listed cases. 

\begin{figure*}
    \centering
    \includegraphics[width=0.8\textwidth]{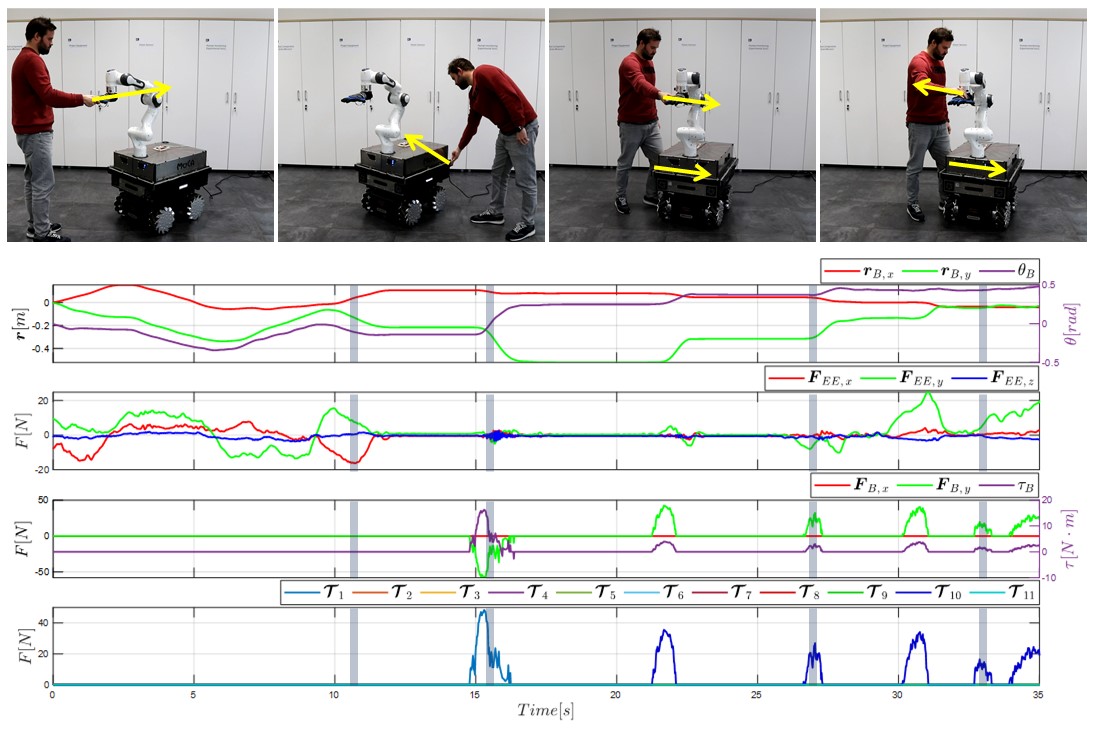}
    \caption{Demonstration of the expanded Follow-me Controller. From top to bottom: i) Screenshots of the experiment, where the yellow arrows define the direction of the applied forces; ii) the pose (position--left axis, orientation--right axis) of the base; iii) the forces applied to the end-effector; iv) the wrenches (force--left axis, and torque--right axis) applied to the base; v) the forces applied to each taxel. The four semi-transparent vertical gray lines represent the moment when the four screenshots represented at the top of the figure were taken.}
    \label{fig:follow_me_demo}
\end{figure*}

\begin{figure*}
    \centering
    \includegraphics[width=0.8\textwidth]{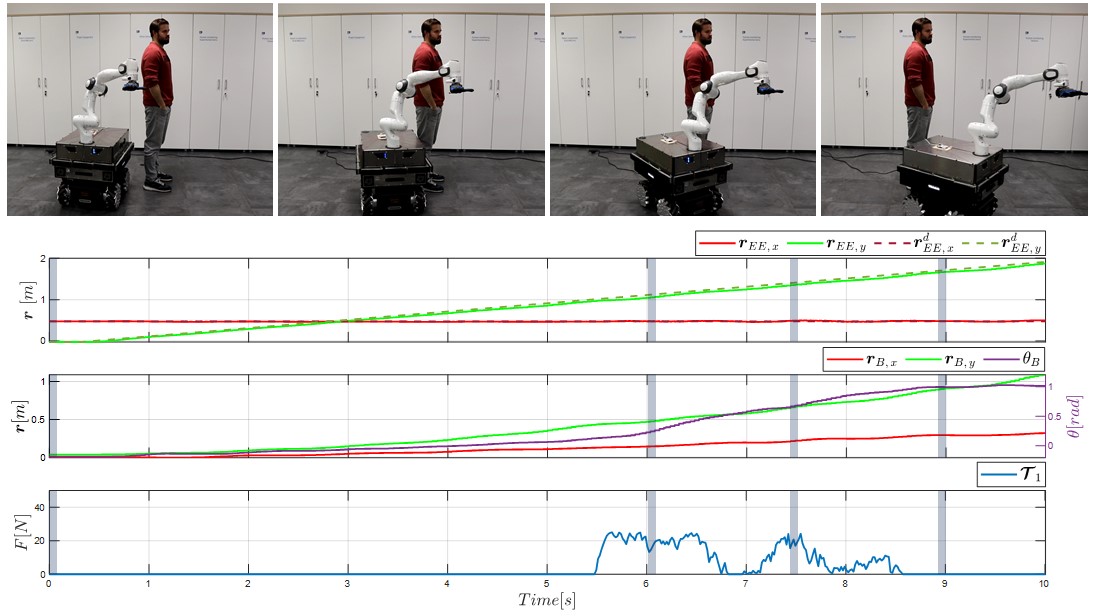}
    \caption{Results of the unexpected collision experiments. From top to bottom: i) Screenshots of the experiment; ii) the end-effector current and desired positions (the Z components are dropped as they are constant during the experiment); ii) the pose (position--left axis, orientation--right axis) of the base; iii) the force applied to the first taxel (the remaining taxels are not considered as they are 0 during the whole experiment). The four semi-transparent vertical gray lines represent the moment when the four screenshots represented at the top of the figure were taken.}
    \label{fig:unexpected_collision_exp}
\end{figure*}

The first case shows how the position of the base (top plot) changes according to the force read at the end-effector (second plot from the top). In the second case, when external forces are applied at the left side ($\mathcal{T}_1$ in the bottom plot), wrenches are read at the base (third plot from the top), causing motions of the base (top plot). In the third case, the same directional forces are simultaneously applied to the end-effector (second plot from the top) and the base ($\mathcal{T}_10$ in the bottom plot), causing a movement of the base (top plot). Two aspects are noticeable in this case: first, as the resulting force is computed in equation~(\ref{eq:virtual_torques_base_follow_me})), a smaller amount of forces causes larger motions of the base; and second, as shown in Fig.~\ref{fig:tactile_sensor}c, the Y component of the end-effector and base frames have opposite directions, explaining why the end-effector forces (second plot from the top) and the base forces (third plot from the top) have opposite sign. On the other hand, in the last case, when opposite forces are applied to the end-effector (second plot from the top) and the base (third plot from the top), the base does not move (top plot). Note that, during the whole experiment, the end-effector configuration remains the same while presenting a compliant behavior, due the inclusion of the arm forces in $\boldsymbol{\tau}_{v,ext}^T$ (second component of equation~(\ref{eq:virtual_torques_base_follow_me})), and the second task (equation~\ref{eq:nullspace_impedance_follow_me}).

\subsection{Unexpected Collision}
\label{subsec:unexpected_collision}

The last experiment consists of forcing an ``unexpected collision'' with an actual human to demonstrate the potential of the MOCA-S platform for ensuring safety in close-contact Human-Robot Collaboration scenarios. In this experiment, the desired trajectory is commanded to the robot's end-effector. This trajectory consists of a displacement in the Y-axis at a constant velocity of $0.2 m/s$ while running the expanded whole-body Cartesian impedance controller. During the motion, the robot's base encounters a human who is not aware of the platform's motion, and a collision occurs. 

The results of the experiment are shown in Fig.~\ref{fig:unexpected_collision_exp}. As in the previous experiments, the vertical gray lines correspond to the four screenshots at the top of the figure. The experiment is analyzed by examining in detail these four particular moments. The platform starts the motion in the first one, following the desired trajectory. Due to the loco-manipulation capabilities introduced by the whole-body controller, the arm moves more than the base at the beginning of the trajectory, then, thanks to the second component of the null-space torque vector in equation~(\ref{eq:nullspace_impedance}), the base follows the end-effector movement. At the second moment, the base collides with the human, and the interaction is measured by $\mathcal{T}_1$ (bottom plot). Note that only this taxel is plotted as it is the only one involved in this experiment. This collision produces a virtual torque $\boldsymbol{\tau}_{v,ext}^T$ and a change of the configuration of the arm w.r.t. the base thanks to the first component of equation~(\ref{eq:nullspace_impedance}), resulting in a safe physical collision with a maximum force of less than $25N$. The bottom plot also exhibits how the base ``bounces'' as revealed by the three peaks of force that decrease in time and magnitude as the collision occurs. The second peak of force corresponds to the third screenshot. The most noticeable aspect is how the orientation of the base changes when the collision takes place (second plot from the top). Finally, the collision is finished, and the robot keeps moving according to the desired trajectory. Moreover, the desired trajectory at the end-effector is correctly tracked even during the collision while the configuration of the arm changes.

\section{Conclusions}
\label{sec:conclusions}
This work addressed the problem of safety in collaborative mobile manipulators. In particular, the problem was tackled by proposing a new Sensitive Mobile Collaborative Robotic Assistant that we called MOCA-S. A low-cost cover made of soft, large-area capacitive tactile sensors was developed and integrated around the platform base to measure the interaction forces applied to the robot base. Four different tactile sensors formed by different materials were evaluated during a compression test. As a result, characterization and calibration of the sensors were carried out. The experiment's outcomes show that the Polyurethane Foam LD30 achieved the best performance. 
Moreover, two expanded whole-body controllers aimed at exploiting the platform's tactile cover and loco-manipulation features were proposed. In particular, an expanded Cartesian impedance controller and Follow-me controllers were implemented. The controllers' performance and robot behavior were evaluated in two experiments. These experiments demonstrated the potential of MOCA-S, which allowed safe physical interaction at the arm and base levels. Finally, a safety experiment was conducted in which an undesired collision between MOCA-S and a human occurs. The outcomes demonstrated the intrinsic safety of the platform. Therefore, the outcomes of this work represent a breakthrough in terms of safety for mobile manipulators.  
Future works will focus on developing more advanced controllers and integrating the cover on different platforms, designing a more robust tactile cover from the mechanical and electrical point of view, and further analysis in a user-study in actual industrial tasks.

\bibliographystyle{ieeetr}
\bibliography{ms.bib}

\end{document}